\documentclass[lettersize, journal]{IEEEtran}
\usepackage{amsmath,amsfonts}
\usepackage{algorithmic}
\usepackage{algorithm}
\usepackage{array}
\usepackage[caption=false,font=normalsize,labelfont=sf,textfont=sf]{subfig}
\usepackage{textcomp}
\usepackage{stfloats}
\usepackage{url}
\usepackage{verbatim}
\usepackage{graphicx}
\usepackage{multirow}
\usepackage{booktabs}
\usepackage{amsmath}
\usepackage{cite}
\usepackage{bm}
\usepackage{xcolor}
\usepackage{setspace}
\usepackage{adjustbox}
\begin{document}

\title{\textbf{MoRE}: \textbf{M}ixture of \textbf{R}esidual \textbf{E}xperts \\ for Humanoid Lifelike Gaits Learning on Complex Terrains}

\author{
\IEEEauthorblockN{
Dewei Wang\textsuperscript{1,2},
Xinmiao Wang\textsuperscript{2,3},
Xinzhe Liu\textsuperscript{2,4},
Jiyuan Shi\textsuperscript{2},
Yingnan Zhao\textsuperscript{3},
Chenjia Bai\textsuperscript{2}\IEEEauthorrefmark{1},
Xuelong Li\textsuperscript{1,2}\IEEEauthorrefmark{1},
}
\IEEEauthorblockA{
\textsuperscript{1}University of Science and Technology of China\\
}
\IEEEauthorblockA{
\textsuperscript{2}Institute of Artificial Intelligence (TeleAI), China Telecom\\
}
\IEEEauthorblockA{
\textsuperscript{3}Harbin Engineering University
}
\IEEEauthorblockA{
\textsuperscript{4}ShanghaiTech University\\
}
\IEEEauthorrefmark{1}Corresponding author.
}



\onecolumn  
\twocolumn[\begin{@twocolumnfalse}
    \maketitle
    \vspace{-3.5em}
    \begin{center}
    {\small \textbf{Website:} \textcolor{orange}{\texttt{\url{https://more-humanoid.github.io/}}}}
    \vspace{2em}
\end{center}
\end{@twocolumnfalse}]



\begin{abstract}
Humanoid robots have demonstrated robust locomotion capabilities using Reinforcement Learning (RL)-based approaches. Further, to obtain human-like behaviors, existing methods integrate human motion-tracking or motion prior in the RL framework. However, these methods are limited in flat terrains with proprioception only, restricting their abilities to traverse challenging terrains with human-like gaits. In this work, we propose a novel framework using a mixture of latent residual experts with multi-discriminators to train an RL policy, which is capable of traversing complex terrains in controllable lifelike gaits with exteroception. Our two-stage training pipeline first teaches the policy to traverse complex terrains using a depth camera, and then enables gait-commanded switching between human-like gait patterns. We also design gait rewards to adjust human-like behaviors like robot base height. Simulation and real-world experiments demonstrate that our framework exhibits exceptional performance in traversing complex terrains, and achieves seamless transitions between multiple human-like gait patterns. 

\end{abstract}


\section{Introduction}

Legged robots have experienced remarkable progress in recent years \cite{dogonchallenge, luo2024pie}. With the development of hardware and control algorithms, humanoid robots have driven more research attention due to their anthropomorphic morphology, which enables them to perform human-like tasks more effectively. Locomotion is one of the primary skills for humanoid robots and serves as a foundation for their applications in various scenarios. With Reinforcement Learning (RL) algorithms, humanoid robots can perform robust locomotion on challenging terrains only with proprioception \cite{loco-wmr, loco1}. Equipped with exteroceptive sensors such as LiDAR and RGB-D cameras, humanoid robots can perform more complex locomotion tasks \cite{pim, humanpark, wang2025beamdojo}. However, the robots often lack anthropomorphism and diversity in their gait behaviors.

\begin{figure}[t]
\centering
\includegraphics[width=1.0\linewidth]{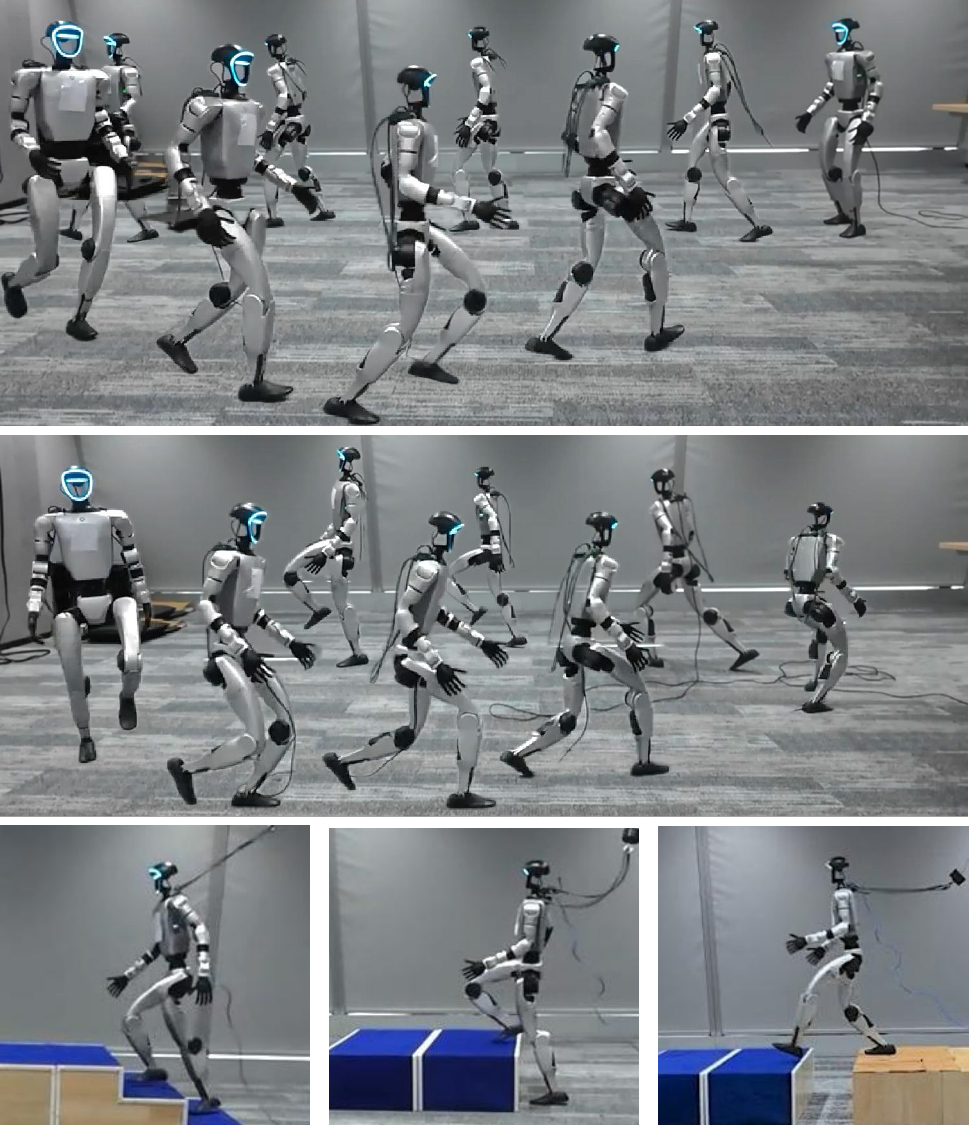}
\caption{Our framework leverages a two-stage training pipeline and the mixture of latent residual experts to enable the humanoid robot to traverse complex terrain with controllable anthropomorphic gaits including walk, run, crouch-walking and high-knees.}
\label{fig:performance}
\end{figure}

Some works for humanoid robots focus on learning human-like behaviors by imitating human motions, where motions can be recorded by a Motion-Capture (MoCap) system or sampled from a motion dataset (e.g., AMASS \cite{mahmood2019amass}). Specifically, this kind of approach trains a policy to encourage the robot to track human motions step-by-step in simulation, and then leverages regularization rewards and domain randomization techniques to perform sim-to-real transfer effectively, enabling them to track smooth \cite{omnih2o, exbody2} or agile motions \cite{asap}. Another way to obtain human-like behavior is through Adversarial Motion Prior (AMP) \cite{amp}, a method that leverages the overall style of motion trajectories to capture natural motion dynamics, rather than stepwise motion tracking \cite{amp4hw, humanmimic}. AMP-based methods demonstrate human-like behavior by injecting knowledge of reference motions into a reward function, which improves the naturalness of locomotion gaits and effectively reduces the complexity of regularization rewards \cite{amp4hw, amp_loco, dogampterrain}. 

However, although the motion-tracking policies trained by imitating the stepwise human behaviors can track complex motions, they often fail to serve as a robust controller for humanoid robot to traverse complex terrains. AMP-based methods can traverse moderately rugged terrain with natural gaits, while just learning from a single reference motion.
Meanwhile, the MoCap data predominantly provides reference motions only for locomotion on flat terrain, making it extremely challenging to leverage such data to learn highly dynamic and balance-critical movements across complex terrains.
Moreover, both motion imitation and AMP-based approaches typically depend solely on proprioception, overlooking the utilization of exteroceptive sensors. This limitation prevents robots from detecting real-time terrain variations, limiting their ability to traverse certain terrains like gaps. Enabling humanoid robots to integrate proprioceptive and exteroceptive sensors to master diverse anthropomorphic gaits for complex terrains is a promising solution.


In this work, we present Mixture of latent Residual Experts (\textbf{MoRE}), aiming to learn multiple human-like gaits that can traverse complex terrains in a single network based on both proprioceptive and exteroceptive sensors. Our framework adopts a two-stage training pipeline: (i) in the first stage, the randomly initialized policy focuses on learning locomotion capability using an exteroceptive sensor without any motion prior; and (ii) in the second stage, we introduce a novel residual module attached to the pretrained policy to learn multiple anthropomorphic gaits, effectively utilizing previously acquired locomotion policy. 
Specifically, the residual module processes multimodal inputs, and incorporates a gait command to explicitly control gait selection. This module outputs a latent feature added to the last hidden layer of the locomotion policy to provide residual information. The motion priors are incorporated through the multiple discriminators in the second stage, where each discriminator is trained on distinct reference motions (as real samples) and robot trajectories (as fake samples) to formulate gait rewards. In policy training, the discriminators are selected based on the gait command to acquire gait-dependent rewards.

Benefiting from the capabilities acquired in the first stage, the final policy combined with the residual module achieves seamless transitions between human-like gaits and robust traversal of complex terrains. To perform multi-gait learning, we propose a Mixture-of-Experts (MoE) \cite{moe, moe_dl}-based architecture for the residual module, which not only accelerates learning but also eliminates gradient conflicts \cite{zhou2022convergence, sodhani2021multi}. The architecture employs a gating network to compute a weighted combination of expert outputs, generating the final latent residual. 
Furthermore, we design specialized gait rewards to achieve finer-grained gait control that can simultaneously learn both reference motions and auxiliary behavioral constraints, facilitating precise gait acquisition instead of being limited by reference motions. Our contributions are summarized as follows:

\begin{itemize}{}{}
    \item \textbf{Two-Stage Paradigm:} We present a two-stage method that employs a single policy to acquire multiple gaits and achieve robust locomotion across complex terrains. 
    \item \textbf{Residual Experts:} We train a mixture of latent residual experts to learn gait-dependent transition with human motion priors from multiple discriminators. 
    \item \textbf{Gait Rewards:} We propose gait-specific rewards for precise behavior control during policy optimization, thereby enhancing the gait diversity accordingly. 
    \item \textbf{Deployment:} The learned policy can be deployed in a real Unitree G1 robot. The experiments exhibit robust locomotion capabilities with multiple human-like gaits. 
\end{itemize}

\section{Related Work}
\subsection{Learning-based Humanoid Locomotion}

Humanoid robots with a deep RL controller trained in highly parallel simulations \cite{isaacgym} exhibit robust and agile locomotion capabilities, where most of this line of research focuses solely on the use of proprioception for locomotion. 
Some previous works \cite{2024realrlloco, loco1, loco2, loco-wmr} utilize proprioceptive information to achieve humanoid robots' robust locomotion on flat or relatively uneven terrains.
ALMI \cite{almi} proposes a novel adversarial training pipeline which iteratively trains both a upper-body policy and a lower-body policy getting a controller capable of resisting multiple disturbances. Drawing inspiration from previous work \cite{walktheseways}, HugWBC \cite{hugwbc} introduces phase variables into the locomotion policy to achieve more diverse gaits. 
Since no exteroceptive sensors are utilized, these methods are unable to fully unleash the robot's potential and traverse complex terrains.

To enable humanoid robots to perceive the environment more directly and comprehensively, some other works integrate LiDAR or depth camera data into the policy. Humanoid parkour with depth camera \cite{humanpark} is achieved by a three-stage training pipeline including policy distillation and an auto-curriculum mechanism. PIM \cite{pim} constructs the elevation map using a LiDAR or RGB-D camera and used contrastive loss for state prediction achieving humanoid locomotion on complex terrain such as stairs and slopes. More challenging locomotion capabilities \cite{vb-com, wang2025beamdojo}, such as walking on sparse footholds and autonomous obstacle avoidance, can be achieved through techniques like multi-policy integration, reward function engineering, and multi-stage training.

\subsection{Anthropomorphic Behavior Learning}
Learning anthropomorphic behavior for humanoid robots has recently garnered significant attention in the research community. By training a policy to perform frame-by-frame tracking of reference motions extracted from videos or MoCap systems, humanoid robots can imitate complex human behaviors, such as backward-leaning, jumping, and shooting \cite{peng2018deepmimic, asap}. OmniH2O \cite{omnih2o} performs motion imitation through the integration of motion re-targeting, feasibility filter, and RL-based policy training, while enabling real-time human motion imitation on robots via a teleoperation system. ExBody2 \cite{exbody2} designes a difficulty level mechanism for reference motion and used motion synthesis to expand motion data, achieving robust and diverse anthropomorphic behavior in humanoid robots. However, these motion tracking approaches fail to enable the humanoid robot to navigate complex terrains.

Using of AMP as a reward function in RL learning has been extensively demonstrated to be an efficient and convenient approach for enabling legged robots to learn animal-level natural gaits \cite{amp, amp4hw, dogampterrain}. Zhang \textit{et al.} \cite{amp_loco} use reference motions retargeted from human demonstrations as motion priors to guide the policy optimization, achieving robust humanoid locomotion on flat and slope terrains. Combing a policy guided by AMP with a safety recovery policy\cite{hwcloco}, the humanoid robot can perform relative natural gait and robust locomotion. The soft-boundary Wasserstein-1 loss with AMP proposed by HumanMimic \cite{humanmimic} helps the policy learn fluent natural locomotion and transitions in simulations.
We also employ AMP for humanoid gait learning. Unlike previous approaches, our policy simultaneously learns multiple gaits as well as gait transitions, with multiple discriminators providing corresponding motion priors.

\begin{figure*}[t]
\centering
\includegraphics[width=1.0\linewidth]{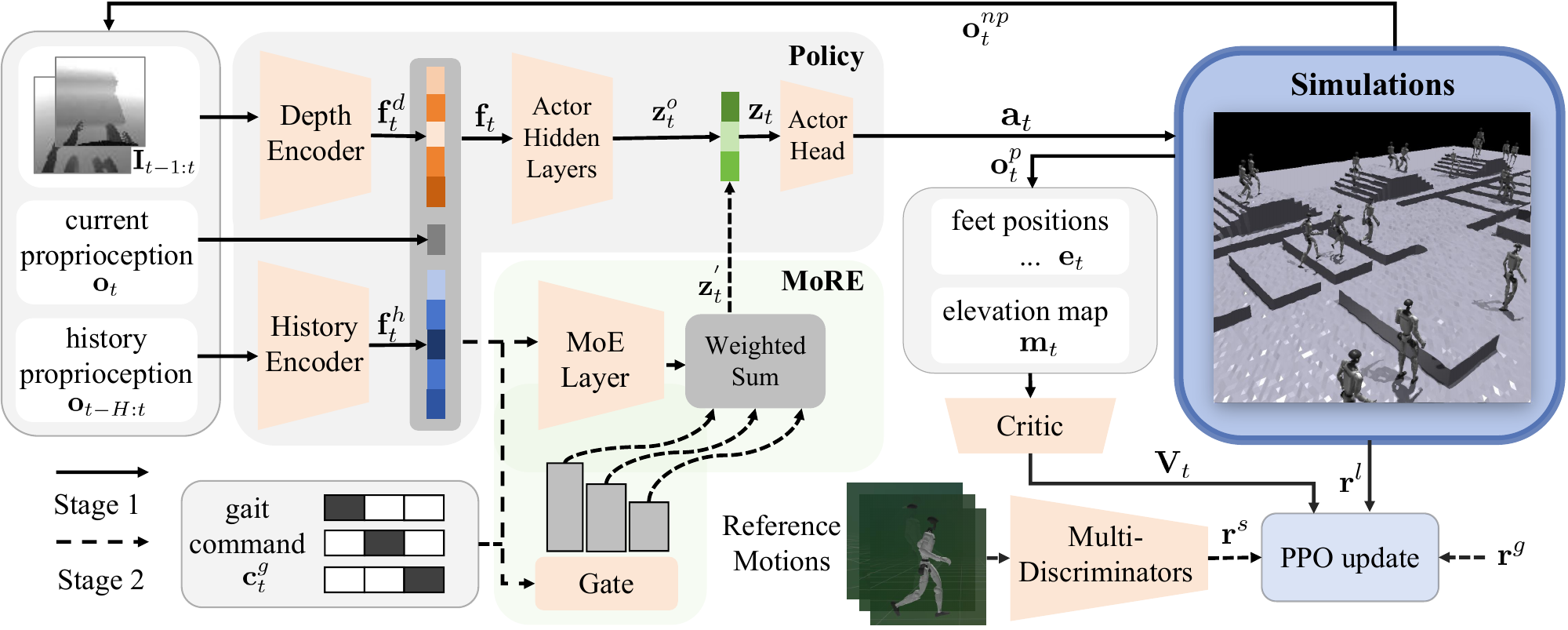}
\caption{Overview of the proposed framework. In the first training stage, we first train a base locomotion policy using only locomotion rewards $\bm{r}^l$ which enables the humanoid robot to traverse complex terrains with a depth camera. In the second training stage, we add a mixture of latent residual experts module to the pretrained base locomotion policy and train them together using multi-discriminators for anthropomorphic gaits learning.}
\label{fig:method_overview}
\end{figure*}

\section{Method}
It is challenging for humanoid robots to traverse complex terrains with both proprioceptive and exteroceptive information while dynamically switching between multiple human-like gaits. This is due to the need for effective comprehension of depth information, robust locomotion across complex terrains, and the learning of diverse anthropomorphic gaits. We address these challenges via a two-stage training pipeline, \emph{MoRE}, which includes locomotion skill learning and anthropomorphic gait acquisition, as illustrated in Fig. \ref{fig:method_overview}. This training pipeline successfully empowers the humanoid robot to traverse complex terrains with gait command-dependent anthropomorphic gaits.
\subsection{Problem Formulation}
We formulate the humanoid locomotion control as a Partially Observable Markov Decision Process (POMDP), defined by a tuple $\mathcal M=(\mathcal S, \mathcal A, \mathcal P, R, \gamma)$ where $\mathcal S $ is the state space, $\mathcal A $ is the action space, $\mathcal P(\cdot|\bm s, \bm a) $ is the transition function, $R: \mathcal S \times \mathcal A \rightarrow \mathcal R$ is the reward function, and $\gamma \in [0,1)$ is the reward discount factor. We adopt the Proximal Policy Optimization (PPO) \cite{schulman2017proximal} for the problem solving with the objective:
\begin{equation}
\label{eq:ppo_obj}
\pi^{*} = \arg\max_{\pi} \mathbb E \big[\sum\nolimits_{t=0}^{T} \gamma^{t}R(\bm{s}_t, \bm{a}_t)\big].
\end{equation}
In the first training stage, the asymmetric actor-critic is applied where the policy takes non-privileged observations $\bm{o}_t^{np} = (\bm{o}_t, \bm{o}_{t-H:t}, \bm{I}_{t-1:t})$ as input where $\bm{o}_t$ is the proprioceptive information including a velocity command and $\bm{I}_t$ represent depth image from camera mounted on the robot head. For better value estimation, the critic takes the privileged observations $\bm{o}_t^{p} = (\bm{m}_t, \bm{e}_t)$ as input, where $\bm{m}_t$ is the elevation map and $\bm{e}_t$ includes privileged information such as feet positions in addition to $\bm{o}_t$ and $ \bm{o}_{t-H:t}$. In addition to their original inputs, the policy and critic receive an additional one-hot encoded gait command $\bm{c}^g_t$ in the second training stage.

\subsection{Base Locomotion Policy}
In the first stage, the actor-critic network is trained from scratch and focus on learning basic locomotion skills to traverse various terrains. The policy is trained using only locomotion rewards $\bm{r}^l$ listed in Table \ref{table:rewards} without any motion prior and gait commands.

We choose the depth camera as the exteroceptive sensor for terrain perception. The proprioception observation $\bm{o}_t$ is defined as:
\begin{equation}
\label{eq:o_t}
\bm{o}_t = [\bm{\omega}_t, \bm{g}_t, \bm{c}^{v}_t, \theta_t, \dot\theta_t, \bm{a}_{t-1}],
\end{equation}
which contains the robot angular velocity $\bm{\omega}_t$, the projected gravity vector $\bm{g}_t$, the velocity command $\bm{c}^{v}_t$ including linear velocity command $\bm{v}_{\text{lin}}^{\text{cmd}}$ and angular velocity command $\bm{\omega}_{\text{yaw}}^{\text{cmd}}$, the last action $\bm{a}_{t-1}$, the joint angle and the joint velocity $\theta_t $ and $\dot\theta_t$. Two consecutive depth images $\bm{I}_{t-1:t}$ and history proprioception $\bm{o}_{t-H:t}$ are encoded by depth encoder and history encoder into depth feature $\bm{f}^d_t$ and history feature $\bm{f}^h_t$. The two encoded feature together with the current proprioception observation $\bm{o}_t$ are concatenated into the actor feature $\bm{f}_t$, which serves as the input of actor hidden layers. The output of actor hidden layers $\bm{z}_t^o$ serves as the input of actor head that predicts the action $\bm{a}_t$ for the robot execution. We divide the actor into two parts because a latent residual $\bm{z}'_t$ will be added to $\bm{z}^o_t$ to obtain the combined latent feature $\bm{z}_t$. Here, $\bm{z}^o_t$ is equal to $\bm{z}_t$. The critic uses an elevation map $\bm{m}_t$ and privileged information $\bm{e}_t$ including robot feet information, ground truth linear velocity and randomized physical parameters in simulations to predict the value $\bm{V}_t$ for policy update. The depth encoder and history encoder adopt the 2D-CNN and 1D-CNN architectures, while the remaining networks are all composed by MLPs.

After this training stage, the policy is capable of traversing stairs, gaps and high platforms, as well as performing robust locomotion in all directions across rough and sloped terrains. 

\subsection{Mixture of Latent Residual Experts}
In the second training stage, we integrate \textbf{MoRE} with the pre-trained base locomotion policy from the previous stage for anthropomorphic gait learning while keeping the original locomotion capabilities. Since learning to traverse complex terrains using natural gaits is quite challenging for humanoid robots, the two-stage training pipeline reduces training complexity while conveniently incorporating the gait command $\bm{c}^g$ to control the gait of the robot.

Residual policy learning\cite{residual, resi_apply} has been proved to be an effective way for improving the performance of a base policy. In our framework, instead of taking the full original observation as base policy and output a residual action, our residual module takes the actor feature $\bm{f}_t$ and the gait command $\bm{c}^g$ as input and outputs a latent residual $\bm{z}'_t$ added to $\bm{z}^o_t$ to serve as input of actor head. The feature possesses a higher density of information. Using such a compact input can decrease the complexity of environmental comprehension of the residual module, thereby reducing the required parameters and enabling the residual module to more efficiently acquire knowledge of the gaits and gait transitions. At the beginning of residual module training, its output is almost totally random and will bring adverse effects to the performance. A suboptimal latent residual induces less detrimental impact than a suboptimal action residual since the pre-trained actor head can ensure that the output actions remain within a reasonable range. Meanwhile, by designing both the input and output to reside in the latent space, the residual module can generate more information-rich outputs to the base policy while eliminating the need to learn the mapping from the latent space to the action space.

To guide the policy to learn gait command-dependent anthropomorphic locomotion, we incorporate the gait command $\bm{c}^g$ into the input of residual module. Motivated by \cite{amp4hw, dogampterrain}, we use AMP for locomotion style learning from reference motions, which designs a discriminator $D_{\phi}$ to predict whether a state transition $(\bm{s}_{t}^{\text{amp}}, \bm{s}_{t+1}^{\text{amp}})$ is sampled from the reference motions or generated by the policy. Here, \(\bm{s}_{t}^{\text{amp}} \in \mathbb{R}^{16}\) consists of the 16 joint angles controlled on the robot. Unlike previous approaches, we replace one state transition with a five-step trajectory $\tau = (\bm{s}_t^{\text{amp}}, \bm{s}_{t+1}^{\text{amp}}, \bm{s}_{t+2}^{\text{amp}}, \bm{s}_{t+3}^{\text{amp}}, \bm{s}_{t+4}^{\text{amp}})$ for a more precise prediction. Since we need the policy to learn more than one distinct anthropomorphic gait simultaneously, we design a multi-discriminator framework, where the objective for the $i$-th discriminator is defined as:
\begin{align}
\label{eq:disc_loss}
\arg\max_{\phi_i}\mathbb{E}_{\tau \sim \mathit{M}_i}&[(D_{\phi_i}(\tau)-1)^2] + \mathbb{E}_{\tau \sim \mathit{G}}[(D_{\phi_i}(\tau)+1)^2] \nonumber \\
& \ \ \ + \frac{\alpha^{d}}{2}\mathbb{E}_{\tau \sim \mathit{M}_i}[\|\nabla_{\phi_i} D_{\phi_i}(\tau)\|_2],
\end{align}
where $\mathit{M}_i$ is the reference motion dataset of $i$-th gait, and $\mathit{G}$ is the dataset generated by the policy interacting with simulations. The first two terms in Eq. \eqref{eq:disc_loss} are least square GAN formulation while the final term is a gradient penalty mitigating the discriminator’s tendency of assigning non-zero gradients on the manifold of the reference data. The style reward of the $i$-th discriminator is computed by:
\begin{align}
\label{eq:style_reward}
 \bm{r}^s_t(\tau) &= \sum_{i=1}^N \text{max}[0, 1-0.25(D_{\phi_i}(\tau)-1)^2] \:\:* \nonumber \\ 
&\ \ \ \ \ \ \ \ \ \ \ \mathbb{I}(\arg\max(\bm{c}^g_t)== i),
\end{align}
which represents that the policy receives style reward only from the output of the corresponding discriminator according to the gait command, and $N$ is the number of gaits. Through the multi-discriminators, the policy can learn multiple gaits as well as transitions based on the reward associated with the gait command. 

Although our two-stage training pipeline reduces training complexity, learning human-like gaits for complex terrain traversal remains challenging.
Traversing different terrains and performing different gaits are both considered as a multi-skill learning task, while combining them makes it extremely hard for a single policy to learn. We thus construct the residual module as an MoE architecture, aiming to utilize a gating mechanism to make a single expert learn similar skills, thus eliminating the gradient conflict problem in direct multi-task policy optimization \cite{gradient1,gradient2}. The final residual module includes $N$ experts and a gate network, both constructed by MLP and taking the same input. The output of residual module $\bm{z}'$ is obtained by performing a weighted sum of the outputs from all experts layers, where the weights are determined by the gate network. It can be defined as:
\begin{equation}
\label{eq:moe_eq}
\bm{z}' = \sum\nolimits_{i=1}^N \bm{z}^e_i \cdot \text{softmax}(\bm{w})[i].
\end{equation}
where $\bm{z}^e_i$ is the output of the $i$-th expert and $\bm{w}$ is a weight predicted by the gate network. The MoE architecture can effectively handle this multi-skill task by activating different experts for different skills. 

\subsection{Gait Rewards}
The policy trained by the proposed two-stage training pipeline is able to perform anthropomorphic gaits on complex terrains, but the locomotion style is strictly constrained by the reference motions. However, if there are shortcomings in the reference motion or the style is not satisfied with our requirement, retraining becomes necessary. To address this problem, we introduce gait rewards, which, similar to the style reward, is specifically designed for different gaits based on the gait command. The gait rewards incorporate constraints on the base height during the crouch-walking, restrictions on leg lift height in the high-knees gait and so on. Through such hand-crafted gait command-dependent rewards, the policy can learn more diverse and desirable natural behaviors rather than strictly replicating the reference motion.

\begin{table}[!t]
\label{table:rewards}
\centering 
\caption{Reward Functions Used in Both Training Stages}
\centering
\renewcommand{\arraystretch}{1.3}
\resizebox{\columnwidth}{!}{
\begin{tabular}{l c l}
\toprule
\textbf{Component} & \textbf{Equation} & \textbf{Weight}\\
\hline
\noalign{\vskip 0.1cm}
\multicolumn{3}{c}{\textbf{Locomotion $\bm{r}^l$}} \\ 
Track lin. vel. & exp$\{ -\frac{||\bm{v}_{\text{lin}}^{\text{cmd}} - \bm{v}_{\text{lin}}||^2_2}{0.25} \}$  & $2.0$ \\
\noalign{\vskip 0.1cm}
Track ang. vel. & exp$\{ -\frac{(\bm{\omega}_{\text{yaw}}^{\text{cmd}} - \bm{\omega}_{\text{yaw}})^2}{0.25} \}$ & $2.0$ \\
Joint acc. & $||\ddot{\theta}||_2^2$ & $-5\text{e}-7$ \\
Joint vel. & $||\dot{\theta}||_2^2$ & $-1\text{e}-3$ \\
Action rate & $||\bm{a}_t - \bm{a}_{t-1}||_2^2$ & $-0.03$ \\
Action smoothness & $||\bm{a}_t - 2\bm{a}_{t-1}+\bm{a}_{t-2}||_2^2$ & $-0.05$ \\
Angular vel. ($x y$)& $||\omega_{xy}||_2^2$ & $-0.05$ \\
Joint power & $|\tau||\dot{ \theta}|^{T}$ & $-2.5\text{e}-5$ \\
Feet stumble & $\mathbb{I}(\exists i, |\bm{F}_i^{xy}| \ge 3|F_i^z|)$ & $-1.0$ \\
Arm deviations & $\displaystyle\sum_{\text{arm joints}}|\theta_i - \theta_{\text{default}}|$ & $-0.5$ \\
Joint pos. limits & $\displaystyle\sum_{\text{all joints}}\bm{out}_i$ & $-2.0$ \\
Joint vel. limits & $RELU(\dot \theta - \dot \theta^{\text{max}})$ & $-1.0$ \\
Torque limits & $RELU(\tau - \tau^{\text{max}})$ & $-1.0$ \\
Feet lateral dist. & $|y^{\text{base}}_i-y^{\text{base}}_j| - d_{\text{min}}$ & $0.5$ \\
Feet slippage & $\displaystyle\sum_{\text{feet}}|\bm{v}_i^{\text{foot}}| * \mathbb{I}_{\text{contact}}$ & $-0.25$ \\
Feet force & $\displaystyle\sum_{\text{feet}}RELU(F_i^z-F_{\text{min}}^{\text{force}})$ & $-2.5\text{e}-4$ \\
Collision & $n_{\text{collision}}$ & $-15.0$ \\
Stuck & $(||\bm{v}||_2\le0.1)*(||\bm{c}^v||_2 \ge 0.2)$ & $-1.0$ \\
Cheat & $\mathbb{I}(|\theta_{\text{heading}}|>1.0)$ & $-2.0$ \\
$y$ axis offset & $|y_{\text{robot}} - y_{\text{start}}|$ & $-2.0$ \\
\hline
\noalign{\vskip 0.1cm}
\multirow{2}{*}{\textbf{Style $\bm{r}^s$}} & $\displaystyle\sum_{i=1}^N \text{max}[0, 1-\frac{1}{4}(D_{\phi_i}(\tau)-1)^2]$ & \multirow{2}{*}{$5.0$} \\
& $* \nonumber \mathbb{I}(\arg\max(\bm{c}^g_t)== i)$ & \\
\noalign{\vskip 0.1cm}
\hline
\multicolumn{3}{c}{\textbf{Gait $\bm{r}^g$}} \\ 
Knee height& $\text{exp}\{ - \frac{|h_{\text{knee}}^{\text{target}} \ - \ \text{max}(\bm{h}_{\text{knees}}^{\text{robot}})|}{0.25} \}$ & $2.0$ \\
\noalign{\vskip 0.1cm}
Squat height& $||h_{\text{squat}}^{\text{target}} - h^{\text{robot}}||_2^2$ & $2.0$ \\
\bottomrule
\end{tabular}}
\end{table}

\section{Training}
\subsubsection{Training Details}The whole training process is performed in the NVIDIA Isaac Gym simulation and uses one NVIDIA RTX 4090 in the first training stage for around 10000 iterations and four NVIDIA RTX 4090s in the second training stage for around 20000 iterations. To accelerate the training process, we use NVIDIA Warp for depth image rendering instead of the original cameras in Isaac Gym. The policy predicts 16-dimensional action $\bm{a}_t \in \mathbb{R}^{16}$ which controls shoulder pitch joints, elbow pitch joints and all leg joints, since the anthropomorphic gaits that the policy to learn is independent of other joints. We use LAFAN1 dataset \cite{harvey2020robust} retargeted to Unitree G1 robot as real samples for training the discriminators.

\subsubsection{Terrain Curriculum}
Terrains used for training in simulations include stairs, gaps, steps and roughness terrains. Since directly learning locomotion on complex terrains is difficult, we adopt an auto-curriculum mechanism that progressively increases terrain difficulty based on policy performance, following the previous work \cite{rudin2022learning}. The terrain curriculum is applied in both stages of training. In the curriculum terrains, the gap width ranges from 0.05$m$ to 0.45$m$, the step height range from 0.05$m$ to 0.3$m$ and the stair height ranges from 0.05$m$ to 0.15$m$. The policy will be moved to harder terrains if it performs well, and to easier terrains if it performs poorly.

\subsubsection{Domain Randomization \& Rewards}
For robust locomotion and better sim-to-real transfer performance, we randomize physics parameters in simulations including the friction coefficient, the restitution coefficient, the mass payload, the center of mass of the robot, the initial joint positions, the motor strength, the PD gains, the action delay and the mass of each link. 
For the ability of resisting external disturbances, we applied external forces to the robot at 8$s$ intervals. To bridge the depth camera gap between the simulation and real world, we add domain randomization and noise to the camera in simulations. We add gaussian noise, depth deviation noise and adapt gaussian filter to the depth image and also randomize the camera position and rotation. Details of domain randomization parameters and their range are listed in Table \ref{table:dr}. Noticing that the peripheral regions in images captured by the head-mounted camera provide no benefit for locomotion tasks, we crop the central region of the image as the policy input. The NVIDIA Warp camera can not render occlusions caused by the robot body and humanoid robots cannot avoid occlusions of their own bodies on cameras in the real world. To solve this issue, we collect some robot body occlusions rendered by Isaac Gym's camera when the second training stage has nearly converged and randomly apply occlusions to the images from NVIDIA Warp camera in the subsequent training allowing the policy to adapt to real-world cameras.

\begin{table}[!t]
\label{table:dr}
\centering 
\caption{Domain Randomization Range of Dynamic Parameters and Depth Camera During Training}
\renewcommand{\arraystretch}{1.2}
\resizebox{\columnwidth}{!}{
\begin{tabular}{l l l}
\hline
\textbf{Term} & \textbf{Randomization Range} & \textbf{Unit}\\
\hline
Friction Coefficient & [0.5, 2.0] & $-$ \\
Mass Payload & [-3.0, 3.0] & Kg\\
Center of Mass Shift & [-3.0, 3.0] & cm \\
Motor Strength Factor & [0.8, 1.2]$\times$motor toeque & Nm\\
Motor Strength Noise & [-0.1, 0.1] & Nm\\
$K_p$ Gains & [0.8, 1.2]$\times$standard value & Nm/rad\\
$K_d$ Gains & [0.8, 1.2]$\times$standard value & Nms/rad\\
Initial Joint Positions & [0.5, 1.5]$\times$nominal value & rad\\
Ground Restitution & [0, 1.0] & $-$\\
Actions Delay & [0, 40] & ms\\
Mass of Each Link & [0.8, 1.2]$\times$nominal value & Kg\\
\hline
Depth Gaussian Noise & [0, 5] & cm\\
Depth Gaussian Filter Kernel & [1, 3, 5] & $-$\\
Depth Gaussian Filter Sigma & [1.2, 1.2] & $-$\\
Depth Deviation Noise & [0, 15] & cm\\
Depth Delay & [0, 8] & ms \\
Camera Position & [-1, 1] & cm\\
Camera Pitch & [42, 48] & deg\\
Camera Horizontal FOV & [77, 81] & deg\\
\hline
\end{tabular}}
\end{table}

Our reward functions can be decomposed into three components: locomotion rewards $\bm{r}^l$, style rewards $\bm{r}^s$ and gait rewards $\bm{r}^g$. In the first training stage, only $\bm{r}^l$ is utilized, whereas the second stage incorporated all reward components. It should be noted that $\bm{r}^s$ and $\bm{r}^g$ are specifically designed to be provided only when their corresponding gait commands are sampled. Table \ref{table:rewards} lists the details of reward functions and their weights.

\section{Results and Discussion}
In this section, we systematically investigate the performance of the proposed framework across complex terrains, the knowledge acquired by individual experts in the \textbf{MoRE} module and the rationale behind applying residuals to the latent space. Meanwhile, our policy can be directly deployed on the real-world robot and perform strong robustness.

We use the Unitree G1 humanoid robot which is equipped with an Intel RealSense D435i for training and real-world deployment. In the real-world experiments, we employ TCP multiprocess communication to achieve robot control and image acquisition. The raw 640 $\times$ 480 resolution images captured from the depth camera are processed by built-in filters in the Intel RealSense API to bridge the gap between simulations and real-world environments and subsequently downsampled to 64×64 resolution as policy inputs. The camera operates at 10 Hz, while the policy runs at 50 Hz. The policy outputs target joint positions, which are then converted into torques via a PD controller to actuate the motors.

\begin{table*}[t]

\centering
\caption{Locomotion performance comparison on multiple terrains, evaluated by success rate (Succ.) and traverse distance (Dist.).}
\resizebox{2\columnwidth}{!}{
\begin{tabular}{ll|cc|cc|cc|cc|cc|cc}
\toprule
\multicolumn{2}{c|}{\multirow{2}{*}{\textbf{Method}}} & 
\multicolumn{2}{c|}{\textbf{Gap} (Easy)} & 
\multicolumn{2}{c|}{\textbf{Gap} (Hard)} &
\multicolumn{2}{c|}{\textbf{Stair} (Easy)} & 
\multicolumn{2}{c|}{\textbf{Stair} (Hard)} &
\multicolumn{2}{c|}{\textbf{Step} (Easy)} & 
\multicolumn{2}{c}{\textbf{Step} (Hard)} \\
\cmidrule{3-14}
 & & Succ. & Dist. & Succ. & Dist. & Succ. & Dist. & Succ. & Dist. & Succ. & Dist. & Succ. & Dist. \\
\midrule

\multicolumn{2}{c|}{Blind Locomotion} & 0.585 & 8.021 & 0.517 & 7.046 & 0.697 & 9.635 & 0.218 & 2.751 & 0.609 & 8.377 & 0.289 & 3.763 \\
\multicolumn{2}{c|}{Base Locomotion} & 0.943 & 13.186 & 0.893 & 12.465 & 0.853 & 11.893 & 0.660 & 9.104 & 0.945 & 13.214 & 0.663 & 9.154 \\

\midrule
\multirow{3}{*}{\textbf{MoRE}} 
 & Walk-Run Gait & \textbf{0.999} & \textbf{13.993} & \textbf{0.933} & \textbf{13.041} & 0.997 & 13.964 & 0.682 & 9.420 & \textbf{1.000} & \textbf{14.000} & 0.777 & 10.796 \\
 & High-Knees Gait & 0.995 & 13.935 & 0.904 & 12.624 & 0.971 & 13.592 & \textbf{0.903} & \textbf{12.606} & 0.989 & 13.849 & \textbf{0.947} & \textbf{13.247} \\
 & Squat Gait & 0.998 & 13.979 & 0.912 & 12.746 & \textbf{0.999} & \textbf{13.992} & 0.816 & 11.357 & \textbf{1.000} & \textbf{14.000} & 0.809 & 11.252 \\

\bottomrule
\label{table:locomotion}
\end{tabular}}
\end{table*}

\subsection{Simulation Experiments}
\label{exp:sim_exp}

\subsubsection{Locomotion Performance}
We evaluate our proposed \textbf{MoRE} against following baseline methods:
(1) \textbf{Blind Locomotion}: trained under the same settings as the first-stage base policy, but without visual inputs. (2) \textbf{Base Locomotion}: The locomotion policy obtained from the first training stage, which lacks lifelike gait adaptation capabilities. In this comparative study, \textbf{MoRE} is implemented with the number of experts set to 3, which is selected based on best practices observed in our experiments.

To evaluate the locomotion capabilities of policies under different terrain conditions, we design a standardized benchmark terrain of size 8$m\times$14$m$. Each track contains one type of obstacle chosen from three categories: gaps, stairs, and steps. For each obstacle type, we define two difficulty levels--Easy and Hard. For gap, the spacing ranges from 0.25-0.4$m$ in Easy mode and 0.4-0.6$m$ in Hard mode. For step and stair, the obstacle height varies from 0.15-0.25$m$ (step) and 0.05-0.15$m$ (stair) in Easy mode, and increases to 0.25-0.35$m$ (step) and 0.15-0.25$m$(stair) in Hard mode. 

We assess locomotion performance through three quantitative measures: (1) 
 \textbf{Success Rate (Succ.)}: The percentage of trails in which the robot successfully reaches the 14$m$ goal within 40$s$ without triggering termination; (2)\textbf{Traversing Distance (Dist.)}: The average distance the robot travels before termination, computed across all trials including both successful and failed attempts.

The results in Table \ref{table:locomotion} clearly demonstrate the advantages of our proposed \textbf{MoRE} framework over baseline methods across a range of terrains. Besides, the results underscore the critical role of visual perception in terrain-aware locomotion, as the blind policy exhibits significantly poorer performance across all terrain types compared to those with visual input. Moreover, incorporating human motion priors substantially enhances generalization to unseen and complex terrains (i.e., Hard mode). A further key strength of \textbf{MoRE} lies in its ability to learn and leverage a diverse set of gait strategies, each tailored to specific terrain challenges. For example, the High-Knees gait is particularly effective on terrains with tall steps or stair-like structures due to its enhanced leg-lifting capability. The Walk-Run gait is well-suited for traversing wide gaps, as its increased forward momentum and stride length enable the robot to bridge discontinuities more effectively. 
By equipping the policy with multiple specialized motion strategies, \textbf{MoRE} significantly enhances adaptability, robustness, and performance across a wide range of environments.

\subsubsection{Policy Component Ablations}
\begin{figure}[t]
\centering
\includegraphics[width=1.0\linewidth]{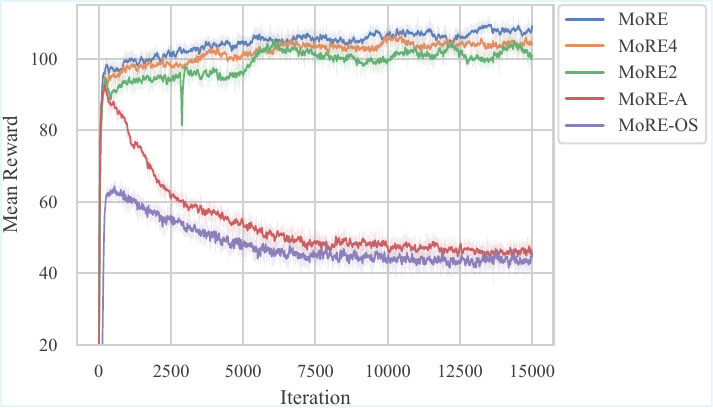}
\caption{The training reward curves under different ablation settings of \textbf{MoRE}.}
\label{fig:ablations}
\end{figure}

To better understand the contribution of each component in our proposed mixture of latent residual experts framework, we conduct several ablation experiments. Specifically, we evaluate the impact of the number of experts, the choice of residual fusion dimension, 
and the initialization strategy. The effectiveness of each variant is assessed based on the training mean episode reward curves. All experiments are performed under consistent settings, and the comparative results are illustrated in Figure~\ref{fig:ablations}.

\textbf{Expert Number}: We perform ablation experiments with expert numbers set to 2, 3 and 4, which correspond to \textit{MoRE2}, \textit{MoRE}, and \textit{MoRE4} in the legend of Figure~\ref{fig:ablations}, respectively. During training, we use tree types of reference gaits: walk\&run, high-knees and squat gait. When using two experts, the high-knees and squat gaits are each associated with a dedicated expert, while walk\&run behavior emerges as a linear combination of the two. With three experts, each expert captures a distinct gait pattern. This leads to improved modularity and more efficient gait composition, resulting in the highest training performance. While, using four experts leads to a decrease in performance. We observe that the high-knees gait is represented by two separate experts, with different expert weightings between the left and right legs. This suggests that the model overfits to minor differences in the reference motion between legs, thereby reducing generalization.

\textbf{Residual Fusion Dimension}: We compare \textbf{MoRE} with a variant that applies residual integration directly in the action space (denoted as \textit{MoRE-A} in Figure~\ref{fig:ablations}). In this variant, the residual network predicts delta actions that are added directly to the output of the base policy which is consistent with common residual policy approaches \cite{asap, resi_apply}. Empirically, this approach results in unstable training and fails to converge, indicating that latent-space fusion provides better gradient flow and more structured modulation of motion features.

\textbf{Policy Initialization Strategy}: Instead of initializing from a pretrained base locomotion policy, we attempt to directly train the residual policy from scratch (one-stage training). This setting leads to complete training failure, as evidenced by the training curve labeled \textit{MoRE-OS} in Figure~\ref{fig:ablations}. These results confirm that a strong base policy provides essential locomotion priors, allowing the residual module to focus on motion specialization and composition. This also highlights the importance of the two stage training scheme.

\begin{figure}[h]
\centering
\includegraphics[width=1.0\linewidth]{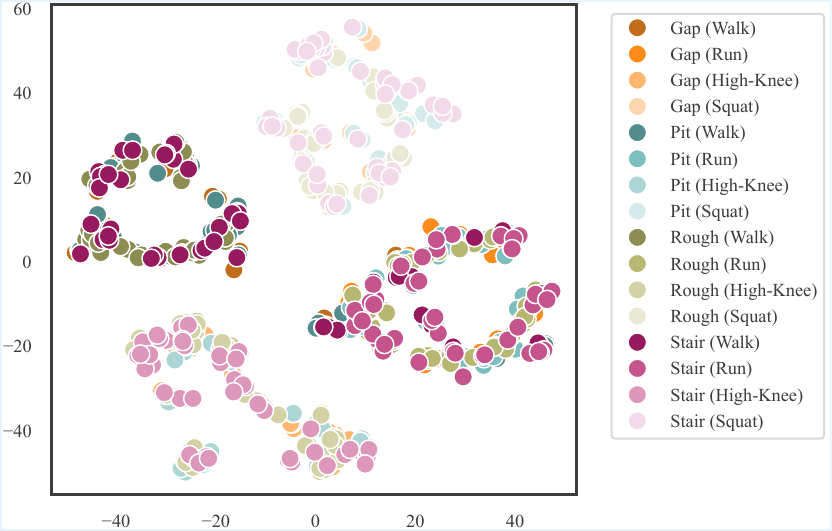}
\caption{The t-SNE visualization of residual latent space across different gaits and terrains.}
\label{fig:tSNE}
\end{figure}

\subsubsection{Residual Latent Analysis}
To further interpret the latent representation learned by each expert, we extract the residual latent outputs from policies trained with three experts and apply t-SNE to project them into a 2D space. As shown in Figure~\ref{fig:tSNE}, samples associated with the same gait form tight clusters, even when collected across varying terrains. This clustering behavior indicates that the residual latent space is semantically structured, capturing high-level motion characteristics rather than overfitting to terrain-specific variations. Furthermore, when a single expert is tasked with handling both walk and run reference motions, the resulting latent vectors form two distinct clusters. This suggests that a single expert is capable of encoding multiple similar motion modes via separable latent features. In addition, we observe that even when executing the walk gait, the residual latent outputs may cluster in the region associated with the run gait during transitions over challenging terrain, such as stair climbing or gap crossing. This suggests that the latent space is not merely reflecting the reference gait labels, but is instead dynamically modulated based on locomotion demands.

\subsubsection{Gait Reward Modulation}
To evaluate the effectiveness of gait reward in modulating motion characteristics, we conduct a series of experiments with different gait-specific targets. We set target values for attributes such as squat height and knee lift height in squat, high-knees gait. As shown in Table \ref{table:gait_reward}, we present the mean achieved values alongside their corresponding target values and the original reference motion values for each gait type. The results demonstrate that the gait reward enables tune specific motion features in a interpretable manner.

\begin{table}[h]
\centering
\caption{Quantitative Evaluation of Gait Reward Modulation}
\resizebox{\columnwidth}{!}{
\begin{tabular}{lccc}
\toprule
\textbf{Attribute Type} & \textbf{Reference} & \textbf{Target} & \textbf{Achieved} \\
\midrule
\multirow{2}{*}{Squat Height (m)} & \multirow{2}{*}{0.680} & 0.600 & 0.669 $\pm$ 0.007 \\
                                  &                       & 0.550 & 0.642 $\pm$ 0.015 \\
\midrule
\multirow{2}{*}{Knee Lift Height (m)} & \multirow{2}{*}{0.922} & 0.580 & 0.612 $\pm$ 0.004 \\
                                     &                       & 0.480 & 0.474 $\pm$ 0.023 \\
\bottomrule
\label{table:gait_reward}
\end{tabular}}
\vspace{-1em}
\end{table}

\subsection{Real-World Experiments}
We deploy the trained \textbf{MoRE} policies onto a Unitree G1 humanoid robot and conduct real-world experiments without any additional fine-tuning, directly transferring the policies from simulation to the robot. To evaluate the generalization and robustness of the proposed method, we test it on several distinct terrain types, including a \textit{Gap} terrain with a 0.4$m$-wide trench, a \textit{Step} terrain with a 0.3$m$ elevation, and a \textit{Stair} terrain composed of three 0.15$m$-high steps. We deploy walk-run, high-knees, and squat gaits on each of these terrains. In addition, we construct a composite terrain that combines the aforementioned obstacles to further evaluate the policy's performance in complex environments. The robot successfully traverses all terrain combinations, even under intentional disturbances (e.g., external pushes), while maintaining stable posture and accurate foot placement. Furthermore, the proposed policy enables smooth and seamless transitions between different gait patterns during traversing terrains. 

\begin{figure}[h]
\centering
\includegraphics[width=1.0\linewidth]{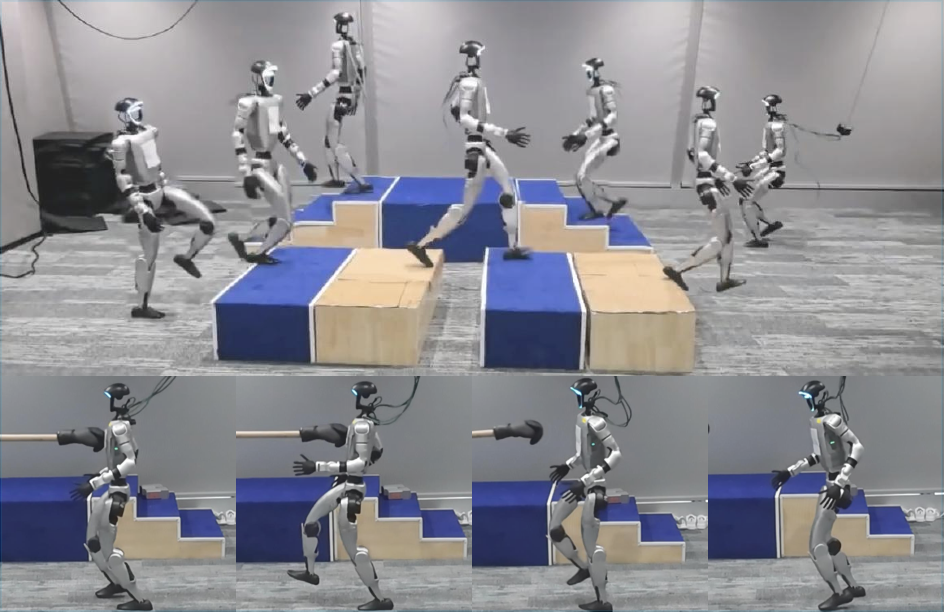}
\caption{Real-world deployment of \textbf{MoRE} on the Unitree G1 humanoid robot. The upper row shows indoor deployment results, where the robot successfully traverses composite terrains.}
\label{fig:real_world_exp}
\end{figure}

As shown in Figure~\ref{fig:real_world_exp}, the robot demonstrates robust and stable behavior across multiple challenging terrain scenarios. Notably, the integration of visual sensing enables the policy to anticipate upcoming terrain changes and generate appropriate responses proactively. This contrasts with prior approaches that rely solely on proprioception, which typically respond reactively to terrain-induced disturbances (e.g., after a collision or misstep). 

\section{Conclusion}
In this work, we proposed a novel framework that integrates visual perception and latent residual experts to enable robust and versatile humanoid locomotion over complex terrains. By leveraging \textbf{MoRE}, our method learns a diverse set of human-like gaits, allowing the robot to reproduce lifelike behaviors. Through extensive simulation and real-world experiments, \textbf{MoRE} demonstrates superior robustness and generalization across a variety of challenging terrains, consistently outperforming baseline policies. In future work, we aim to extend our framework to learn more challenging and lifelike gaits and achieving smoother gait transitions.

\bibliography{main}

\begin{thebibliography}{10}
\providecommand{\url}[1]{#1}
\csname url@samestyle\endcsname
\providecommand{\newblock}{\relax}
\providecommand{\bibinfo}[2]{#2}
\providecommand{\BIBentrySTDinterwordspacing}{\spaceskip=0pt\relax}
\providecommand{\BIBentryALTinterwordstretchfactor}{4}
\providecommand{\BIBentryALTinterwordspacing}{\spaceskip=\fontdimen2\font plus
\BIBentryALTinterwordstretchfactor\fontdimen3\font minus \fontdimen4\font\relax}
\providecommand{\BIBforeignlanguage}[2]{{%
\expandafter\ifx\csname l@#1\endcsname\relax
\typeout{** WARNING: IEEEtran.bst: No hyphenation pattern has been}%
\typeout{** loaded for the language `#1'. Using the pattern for}%
\typeout{** the default language instead.}%
\else
\language=\csname l@#1\endcsname
\fi
#2}}
\providecommand{\BIBdecl}{\relax}
\BIBdecl

\bibitem{dogonchallenge}
J.~Lee, J.~Hwangbo, L.~Wellhausen, V.~Koltun, and M.~Hutter, ``Learning quadrupedal locomotion over challenging terrain,'' \emph{Science robotics}, vol.~5, no.~47, p. eabc5986, 2020.

\bibitem{luo2024pie}
S.~Luo, S.~Li, R.~Yu, Z.~Wang, J.~Wu, and Q.~Zhu, ``Pie: Parkour with implicit-explicit learning framework for legged robots,'' \emph{IEEE Robotics and Automation Letters}, 2024.

\bibitem{loco-wmr}
W.~Sun, L.~Chen, Y.~Su, B.~Cao, Y.~Liu, and Z.~Xie, ``Learning humanoid locomotion with world model reconstruction,'' \emph{arXiv preprint arXiv:2502.16230}, 2025.

\bibitem{loco1}
W.~Cui, S.~Li, H.~Huang, B.~Qin, T.~Zhang, L.~Zheng, Z.~Tang, C.~Hu, N.~Yan, J.~Chen \emph{et~al.}, ``Adapting humanoid locomotion over challenging terrain via two-phase training,'' in \emph{8th Annual Conference on Robot Learning}, 2024.

\bibitem{pim}
J.~Long, J.~Ren, M.~Shi, Z.~Wang, T.~Huang, P.~Luo, and J.~Pang, ``Learning humanoid locomotion with perceptive internal model,'' \emph{arXiv preprint arXiv:2411.14386}, 2024.

\bibitem{humanpark}
Z.~Zhuang, S.~Yao, and H.~Zhao, ``Humanoid parkour learning,'' \emph{arXiv preprint arXiv:2406.10759}, 2024.

\bibitem{wang2025beamdojo}
H.~Wang, Z.~Wang, J.~Ren, Q.~Ben, T.~Huang, W.~Zhang, and J.~Pang, ``Beamdojo: Learning agile humanoid locomotion on sparse footholds,'' in \emph{Robotics: Science and Systems ({RSS})}, 2025.

\bibitem{mahmood2019amass}
N.~Mahmood, N.~Ghorbani, N.~F. Troje, G.~Pons-Moll, and M.~J. Black, ``Amass: Archive of motion capture as surface shapes,'' in \emph{Proceedings of the IEEE/CVF international conference on computer vision}, 2019, pp. 5442--5451.

\bibitem{omnih2o}
T.~He, Z.~Luo, X.~He, W.~Xiao, C.~Zhang, W.~Zhang, K.~Kitani, C.~Liu, and G.~Shi, ``Omnih2o: Universal and dexterous human-to-humanoid whole-body teleoperation and learning,'' \emph{arXiv preprint arXiv:2406.08858}, 2024.

\bibitem{exbody2}
M.~Ji, X.~Peng, F.~Liu, J.~Li, G.~Yang, X.~Cheng, and X.~Wang, ``Exbody2: Advanced expressive humanoid whole-body control,'' \emph{arXiv preprint arXiv:2412.13196}, 2024.

\bibitem{asap}
T.~He, J.~Gao, W.~Xiao, Y.~Zhang, Z.~Wang, J.~Wang, Z.~Luo, G.~He, N.~Sobanbab, C.~Pan \emph{et~al.}, ``Asap: Aligning simulation and real-world physics for learning agile humanoid whole-body skills,'' \emph{arXiv preprint arXiv:2502.01143}, 2025.

\bibitem{amp}
X.~B. Peng, Z.~Ma, P.~Abbeel, S.~Levine, and A.~Kanazawa, ``Amp: Adversarial motion priors for stylized physics-based character control,'' \emph{ACM Transactions on Graphics (ToG)}, vol.~40, no.~4, pp. 1--20, 2021.

\bibitem{amp4hw}
A.~Escontrela, X.~B. Peng, W.~Yu, T.~Zhang, A.~Iscen, K.~Goldberg, and P.~Abbeel, ``Adversarial motion priors make good substitutes for complex reward functions,'' in \emph{2022 IEEE/RSJ International Conference on Intelligent Robots and Systems (IROS)}.\hskip 1em plus 0.5em minus 0.4em\relax IEEE, 2022, pp. 25--32.

\bibitem{humanmimic}
A.~Tang, T.~Hiraoka, N.~Hiraoka, F.~Shi, K.~Kawaharazuka, K.~Kojima, K.~Okada, and M.~Inaba, ``Humanmimic: Learning natural locomotion and transitions for humanoid robot via wasserstein adversarial imitation,'' in \emph{2024 IEEE International Conference on Robotics and Automation (ICRA)}.\hskip 1em plus 0.5em minus 0.4em\relax IEEE, 2024, pp. 13\,107--13\,114.

\bibitem{amp_loco}
Q.~Zhang, P.~Cui, D.~Yan, J.~Sun, Y.~Duan, G.~Han, W.~Zhao, W.~Zhang, Y.~Guo, A.~Zhang \emph{et~al.}, ``Whole-body humanoid robot locomotion with human reference,'' in \emph{2024 IEEE/RSJ International Conference on Intelligent Robots and Systems (IROS)}.\hskip 1em plus 0.5em minus 0.4em\relax IEEE, 2024, pp. 11\,225--11\,231.

\bibitem{dogampterrain}
J.~Wu, G.~Xin, C.~Qi, and Y.~Xue, ``Learning robust and agile legged locomotion using adversarial motion priors,'' \emph{IEEE Robotics and Automation Letters}, vol.~8, no.~8, pp. 4975--4982, 2023.

\bibitem{moe}
R.~A. Jacobs, M.~I. Jordan, S.~J. Nowlan, and G.~E. Hinton, ``Adaptive mixtures of local experts,'' \emph{Neural computation}, vol.~3, no.~1, pp. 79--87, 1991.

\bibitem{moe_dl}
Z.~Chen, Y.~Deng, Y.~Wu, Q.~Gu, and Y.~Li, ``Towards understanding the mixture-of-experts layer in deep learning,'' \emph{Advances in neural information processing systems}, vol.~35, pp. 23\,049--23\,062, 2022.

\bibitem{zhou2022convergence}
S.~Zhou, W.~Zhang, J.~Jiang, W.~Zhong, J.~Gu, and W.~Zhu, ``On the convergence of stochastic multi-objective gradient manipulation and beyond,'' \emph{Advances in Neural Information Processing Systems}, vol.~35, pp. 38\,103--38\,115, 2022.

\bibitem{sodhani2021multi}
S.~Sodhani, A.~Zhang, and J.~Pineau, ``Multi-task reinforcement learning with context-based representations,'' in \emph{International Conference on Machine Learning}.\hskip 1em plus 0.5em minus 0.4em\relax PMLR, 2021, pp. 9767--9779.

\bibitem{isaacgym}
V.~Makoviychuk, L.~Wawrzyniak, Y.~Guo, M.~Lu, K.~Storey, M.~Macklin, D.~Hoeller, N.~Rudin, A.~Allshire, A.~Handa \emph{et~al.}, ``Isaac gym: High performance gpu-based physics simulation for robot learning,'' \emph{arXiv preprint arXiv:2108.10470}, 2021.

\bibitem{2024realrlloco}
I.~Radosavovic, T.~Xiao, B.~Zhang, T.~Darrell, J.~Malik, and K.~Sreenath, ``Real-world humanoid locomotion with reinforcement learning,'' \emph{Science Robotics}, vol.~9, no.~89, p. eadi9579, 2024.

\bibitem{loco2}
X.~Gu, Y.-J. Wang, X.~Zhu, C.~Shi, Y.~Guo, Y.~Liu, and J.~Chen, ``Advancing humanoid locomotion: Mastering challenging terrains with denoising world model learning,'' \emph{arXiv preprint arXiv:2408.14472}, 2024.

\bibitem{almi}
J.~Shi, X.~Liu, D.~Wang, O.~Lu, S.~Schwertfeger, F.~Sun, C.~Bai, and X.~Li, ``Adversarial locomotion and motion imitation for humanoid policy learning,'' \emph{arXiv preprint arXiv:2504.14305}, 2025.

\bibitem{walktheseways}
G.~B. Margolis and P.~Agrawal, ``Walk these ways: Tuning robot control for generalization with multiplicity of behavior,'' in \emph{Conference on Robot Learning}.\hskip 1em plus 0.5em minus 0.4em\relax PMLR, 2023, pp. 22--31.

\bibitem{hugwbc}
Y.~Xue, W.~Dong, M.~Liu, W.~Zhang, and J.~Pang, ``A unified and general humanoid whole-body controller for fine-grained locomotion,'' \emph{arXiv preprint arXiv:2502.03206}, 2025.

\bibitem{vb-com}
J.~Ren, T.~Huang, H.~Wang, Z.~Wang, Q.~Ben, J.~Pang, and P.~Luo, ``Vb-com: Learning vision-blind composite humanoid locomotion against deficient perception,'' \emph{arXiv preprint arXiv:2502.14814}, 2025.

\bibitem{peng2018deepmimic}
X.~B. Peng, P.~Abbeel, S.~Levine, and M.~Van~de Panne, ``Deepmimic: Example-guided deep reinforcement learning of physics-based character skills,'' \emph{ACM Transactions On Graphics (TOG)}, vol.~37, no.~4, pp. 1--14, 2018.

\bibitem{hwcloco}
S.~Lin, G.~Qiao, Y.~Tai, A.~Li, K.~Jia, and G.~Liu, ``Hwc-loco: A hierarchical whole-body control approach to robust humanoid locomotion,'' \emph{arXiv preprint arXiv:2503.00923}, 2025.

\bibitem{schulman2017proximal}
J.~Schulman, F.~Wolski, P.~Dhariwal, A.~Radford, and O.~Klimov, ``Proximal policy optimization algorithms,'' \emph{arXiv preprint arXiv:1707.06347}, 2017.

\bibitem{residual}
T.~Silver, K.~Allen, J.~Tenenbaum, and L.~Kaelbling, ``Residual policy learning,'' \emph{arXiv preprint arXiv:1812.06298}, 2018.

\bibitem{resi_apply}
A.~Zeng, S.~Song, J.~Lee, A.~Rodriguez, and T.~Funkhouser, ``Tossingbot: Learning to throw arbitrary objects with residual physics,'' \emph{IEEE Transactions on Robotics}, vol.~36, no.~4, pp. 1307--1319, 2020.

\bibitem{gradient1}
T.~Yu, S.~Kumar, A.~Gupta, S.~Levine, K.~Hausman, and C.~Finn, ``Gradient surgery for multi-task learning,'' \emph{Advances in neural information processing systems}, vol.~33, pp. 5824--5836, 2020.

\bibitem{gradient2}
B.~Liu, X.~Liu, X.~Jin, P.~Stone, and Q.~Liu, ``Conflict-averse gradient descent for multi-task learning,'' \emph{Advances in Neural Information Processing Systems}, vol.~34, pp. 18\,878--18\,890, 2021.

\bibitem{harvey2020robust}
F.~G. Harvey, M.~Yurick, D.~Nowrouzezahrai, and C.~Pal, ``Robust motion in-betweening,'' vol.~39, no.~4, 2020.

\bibitem{rudin2022learning}
N.~Rudin, D.~Hoeller, P.~Reist, and M.~Hutter, ``Learning to walk in minutes using massively parallel deep reinforcement learning,'' in \emph{Conference on Robot Learning}.\hskip 1em plus 0.5em minus 0.4em\relax PMLR, 2022, pp. 91--100.

\end{thebibliography}
\bibliographystyle{IEEEtran}

\vfill

\end{document}